\documentclass[conference]{IEEEtran}
\ifCLASSINFOpdf
  \usepackage[pdftex]{graphicx}
  \usepackage{amssymb}
  \usepackage{amsmath} 
  \usepackage{booktabs}
  \usepackage[table]{xcolor}
  \usepackage{float}
  \usepackage{placeins}
   \usepackage{subcaption}

  \usepackage{cite}
    \usepackage{xcolor}
    \usepackage{hyperref}
    \hypersetup{
        colorlinks=true,
        citecolor=red
    }

\else
\fi
\UseRawInputEncoding
\begin{document}
\title{HyMAD: A Hybrid Multi-Activity Detection Approach for Border Surveillance and Monitoring}

\author{
    \IEEEauthorblockN{Sriram Srinivasan}
    \IEEEauthorblockA{
        \textit{CSIR-CMERI} \\
    }
    \and
    \IEEEauthorblockN{Srinivasan Aruchamy}
    \IEEEauthorblockA{
        \textit{CSIR-CMERI} \\
    }
    \and 
    \IEEEauthorblockN{Siva Ram Krishna Vadali}
    \IEEEauthorblockA{
        \textit{CSIR-CMERI} \\
    }
}

\maketitle
\begin{abstract}
Seismic sensing has emerged as a promising solution for border surveillance and monitoring; the seismic sensors that are often buried underground are small and cannot be noticed easily, making them difficult for intruders to detect, avoid, or vandalize. This significantly enhances their effectiveness compared to highly visible cameras or fences. However, accurately detecting and distinguishing between overlapping activities that are happening simultaneously, such as human intrusions, animal movements, and vehicle rumbling, remains a major challenge due to the complex and noisy nature of seismic signals. Correctly identifying simultaneous activities is critical because failing to separate them can lead to misclassification, missed detections, and an incomplete understanding of the situation, thereby reducing the reliability of surveillance systems. To tackle this problem, we propose HyMAD (Hybrid Multi-Activity Detection), a deep neural architecture based on spatio-temporal feature fusion. The framework integrates spectral features extracted with SincNet and temporal dependencies modeled by a recurrent neural network (RNN). In addition, HyMAD employs self-attention layers to strengthen intra-modal representations and a cross-modal fusion module to achieve robust multi-label classification of seismic events. e evaluate our approach on a dataset constructed from real-world field recordings collected in the context of border surveillance and monitoring, demonstrating its ability to generalize to complex, simultaneous activity scenarios involving humans, animals, and vehicles. Our method achieves competitive performance and offers a modular framework for extending seismic-based activity recognition in real-world security applications.
\end{abstract}

\IEEEpeerreviewmaketitle
\section{Introduction}
The use of seismic sensor networks has unlocked unprecedented opportunities for monitoring and interpreting activity patterns in various environments ranging from earth's geological movement to border surveillance. Seismic signals, generated by ground vibrations, carry rich information about various sources, including anthropogenic activities (e.g., human footfalls, animal movements, vehicle passages). Accurate and automated detection and classification of these activities from seismic data are very important for border surveillance and monitoring.

Despite significant advances in the classification of isolated activities like a lone person walking or a vehicle passing \cite{10630594} \cite{HASABNIS2025105815}, the robust identification of multiple, simultaneously occurring activities like human and animal movements, or a vehicle passing alongside wildlife remains a central and largely open challenge in seismic signal analysis. Such overlapping signatures present a significant hurdle for traditional signal processing and machine learning approaches, as the combined vibrational patterns can obscure individual components, leading to misclassification or missed
detections. Existing methodologies often struggle with the intricate fusion of diverse signal characteristics necessary to disentangle these complex, multi-source seismic events. Traditional seismic signal analysis has often relied on handcrafted spectral, temporal, or time-frequency features. However, these manually designed features typically struggle to capture the intricate and overlapping patterns that arise when multiple activities are happening simultaneously, which degrades performance and leads to poor generalizability in realistic, complex monitoring scenarios.

Correctly identifying simultaneous activities is particularly important because missing one activity in the presence of another can significantly compromise the reliability of border surveillance systems. Prior studies have shown that overlapping seismic signals complicate detection and classification, often leading to incomplete or inaccurate monitoring records \cite{yoon2017overlap}, while seismic data analysis literature emphasizes that overlapping signals from natural and anthropogenic sources must be effectively decomposed to enable robust interpretation \cite{yilmaz2001seismic}. These findings highlight that resolving multi-source seismic events is not merely a technical improvement but a fundamental requirement for developing reliable security and monitoring systems.

To overcome these limitations in multi-activity seismic event detection, we propose \textbf{HyMAD} (Hybrid Multi-Activity Detection) a novel deep neural framework designed to exploit the complementary strengths of both frequency and temporal features. Our HyMAD architecture initiates processing with a specialized SincNet-based convolutional layer that extracts interpretable, learnable frequency features via parameterized band-pass filters. These features serve as input to a recurrent neural network (RNN), which captures the spectral evolution and sequential dependencies. Both frequency and temporal representations are further refined through a self-attention module of their own, enabling each modality to contextually enhance its own features. Central to our approach is a cross-attention mechanism that adaptively fuses the two feature sequences, facilitating rich inter-domain interaction. The resulting fused representation is classified by a multi-layer perceptron (MLP) to identify one or more simultaneously occurring activities.

The primary contributions of this paper are summarized as follows:
\begin{itemize}
    \item \textbf{Novel HyMAD Architecture:} We introduce a novel hybrid deep learning framework for robust detection of concurrent human, animal, and vehicle activities from raw seismic signals.
    \item \textbf{Principled Frequency Feature Extraction:} Integration of a SincNet's parametrized convolutional layer for learning interpretable and relevant frequency features directly from raw seismic waveforms.
    \item \textbf{Adaptive Multi-Modal Feature Fusion:} A sophisticated attention-based strategy combining RNN-encoded temporal dynamics with SincNet-derived frequency features through self and cross-attention blocks for robust disentanglement of overlapping signatures.
    \item \textbf{Comprehensive Overlapping Activity Evaluation:} Demonstrated superior performance in identifying complex, overlapping activity scenarios through extensive real-world seismic dataset experiments.
\end{itemize}

\section{Related Work}

The field of seismic signal analysis for activity detection has evolved significantly, driven by advancements in sensor technology and machine learning. Our proposed Hybrid Multi-Activity Detection (HyMAD) architecture builds upon foundational work in seismic signal processing, deep learning for time series, and multi-modal feature fusion. This section summarizes existing research on recognizing human, animal, and vehicle activities, including their simultaneous occurrences, and discusses how our method addresses the shortcomings of previous approaches.

\subsection{Traditional Seismic Activity Classification}
Early work on seismic activity classification aimed at distinguishing sources such as human footfalls, animal movements, and vehicle passages primarily relied on classical signal processing techniques. Methods such as Short-Time Fourier Transform (STFT) \cite{1164317}, Wavelet Transform \cite{morlet1982wave}, and various statistical features (e.g., Root Mean Square, Zero-Crossing Rate) were commonly employed to characterize seismic events. These extracted features were then typically fed into conventional machine learning classifiers like Support Vector Machines (SVMs) \cite{cortes1995support}, Random Forests \cite{breiman2001random}, or Hidden Markov Models (HMMs) \cite{rabiner1989tutorial} for activity classification. While these approaches offer a degree of interpretability due to their explicit feature engineering and can perform adequately for simple, isolated events in controlled environments, they are heavily reliant on extensive domain expertise for effective feature selection. This manual feature engineering process is often time-consuming and struggles to generalize across diverse environmental conditions or to novel activity types. More critically, these traditional methods demonstrate limited robustness and accuracy when faced with the inherent variability, noise, and particularly the overlapping signatures of multiple concurrent activities (e.g., human and animal, human and vehicle) prevalent in real-world seismic data. Their constrained capacity for adaptive feature learning and complex pattern recognition often leads to misclassification or missed detections when multiple activities co-occur and their unique features blend.

\subsection{Deep Learning for Seismic Signal Analysis}
The advent of deep learning has significantly advanced time-series analysis, with increasing applications in seismic activity detection. Convolutional Neural Networks (CNNs) have been adopted for their ability to automatically learn hierarchical features directly from raw seismic waveforms, thereby reducing the reliance on manual feature engineering \cite{lecun1998gradient}. Recurrent Neural Networks (RNNs) \cite{elman1990finding}, including Long Short-Term Memory (LSTM) \cite{hochreiter1997long} and Gated Recurrent Unit (GRU) networks \cite{cho2014learning}, have shown promise in capturing temporal dependencies in seismic data for event classification and localization. These deep learning models offer superior capabilities in automatically learning complex, high-level features directly from raw data, generally leading to improved performance over traditional methods. However, a common limitation in many existing deep learning models applied to seismic activity detection is their primary focus on single-event classification (e.g., classifying a signal as \textit{either} human \textit{or} vehicle) or simpler binary detection tasks (e.g., activity vs. no activity). When confronted with complex multi-activity scenarios, these models frequently employ simpler fusion strategies, such as late fusion (concatenating features before a final classifier) or lack explicit mechanisms to effectively disentangle overlapping temporal and spectral information. This inherent limitation restricts their ability to accurately identify and separate concurrent activities (e.g., simultaneously detecting human and animal movements), consequently limiting their performance in complex, multi-source environments where fine-grained, multi-label activity detection is required.

\subsection{Multi-Scale Frequency Feature Learning}
In the analysis of complex signals with overlapping sources, multi-scale feature extraction is a fundamental strategy. By operating at multiple resolutions, models can capture both broad low-frequency patterns and fine, short-lived details, which is essential for accurate signal separation and classification. Ravanelli and Bengio \cite{ravanelli2018speaker} introduced SincNet, a principled approach to frequency feature extraction that explicitly learns frequency-specific components. Instead of using generic convolutional kernels, SincNet parameterizes the first convolutional layer with rectangular band-pass filters, enabling the network to focus on meaningful frequency bands and directly learn interpretable representations from raw waveforms without relying on traditional spectral transformations such as the STFT. Building on this, Chang \textit{et al}. \cite{chang2021mssincresnetjointlearning1d} developed MS-SincResNet for music classification, employing parallel SincNet layers with different filter lengths to form a rich time--frequency representation; short filters capture transient percussive events, while longer filters resolve fine harmonic structures. Extending this concept further, Mayor-Torres \textit{et al}. \cite{mayortorres2021interpretablesincnetbaseddeeplearning} proposed SincNet-R for EEG signal based emotion recognition, which automatically extracts the most discriminative neural frequency bands directly from the data.

In neuroscience, accurate motor imagery EEG recognition across multiple subjects and categories is challenging due to overlapping frequency components. To address this, Luo \textit{et al} \cite{Luo2023} proposed an overlapping filter bank CNN that decomposes EEG signals into distinct frequency bands and extracts discriminative features from each, capturing both high and low-frequency patterns for enhanced robustness and classification performance. Similarly, Chen \textit{et al}. \cite{Chen2023} tackle the problem of classifying sound signals in reverberant environments, where reflections and environmental distortions can obscure the original signal. They employ densely connected networks that leverage multiple feature types within a multi-scale architecture, enabling the model to effectively separate the original sound from its distorted reflections.

\subsection{Multi-Modal Feature Fusion and Attention Mechanisms}
Once multi-scale or multi-modal features are extracted, the way they are fused becomes critical for robust classification. A straightforward strategy is feature concatenation, where multiple feature vectors are joined end-to-end before being passed to a classifier. This approach has been widely adopted and shown to outperform single-feature baselines in tasks such as acoustic scene classification \cite{Fedorishin2021}  \cite{McLoughlin2020} and arrhythmia diagnosis \cite{Wang2022}. While concatenation provides a direct means of combining complementary information, such as time-domain waveforms with frequency-domain spectrograms.

More advanced techniques employ attention mechanisms, which enable models to adaptively weigh the importance of different features which is very important in noisy or complex environments. Jin \textit{et al}. \cite{Jin2022} proposed an attention-based fusion approach for polyphonic sound event detection, where temporal, frequency, and feature-space attention are jointly leveraged to highlight the most relevant cues from overlapping sources. Building on this idea, cross-attention has recently emerged as a state-of-the-art fusion paradigm. Cross-attention provides a flexible way to fuse heterogeneous features by allowing one feature stream to query another and emphasize the most informative components. Huang et al. \cite{202412.1180} applied this idea to acoustic scene classification, where CNN-extracted frequency features were fused with GRU-based temporal embeddings through cross-attention. In their design, the temporal representation serves as the query to selectively highlight salient local spectral details, enabling an adaptive integration of time and frequency information for improved classification accuracy.

\section{Method}

\begin{figure*}[!t]
    \centering
    \includegraphics[width=\textwidth]{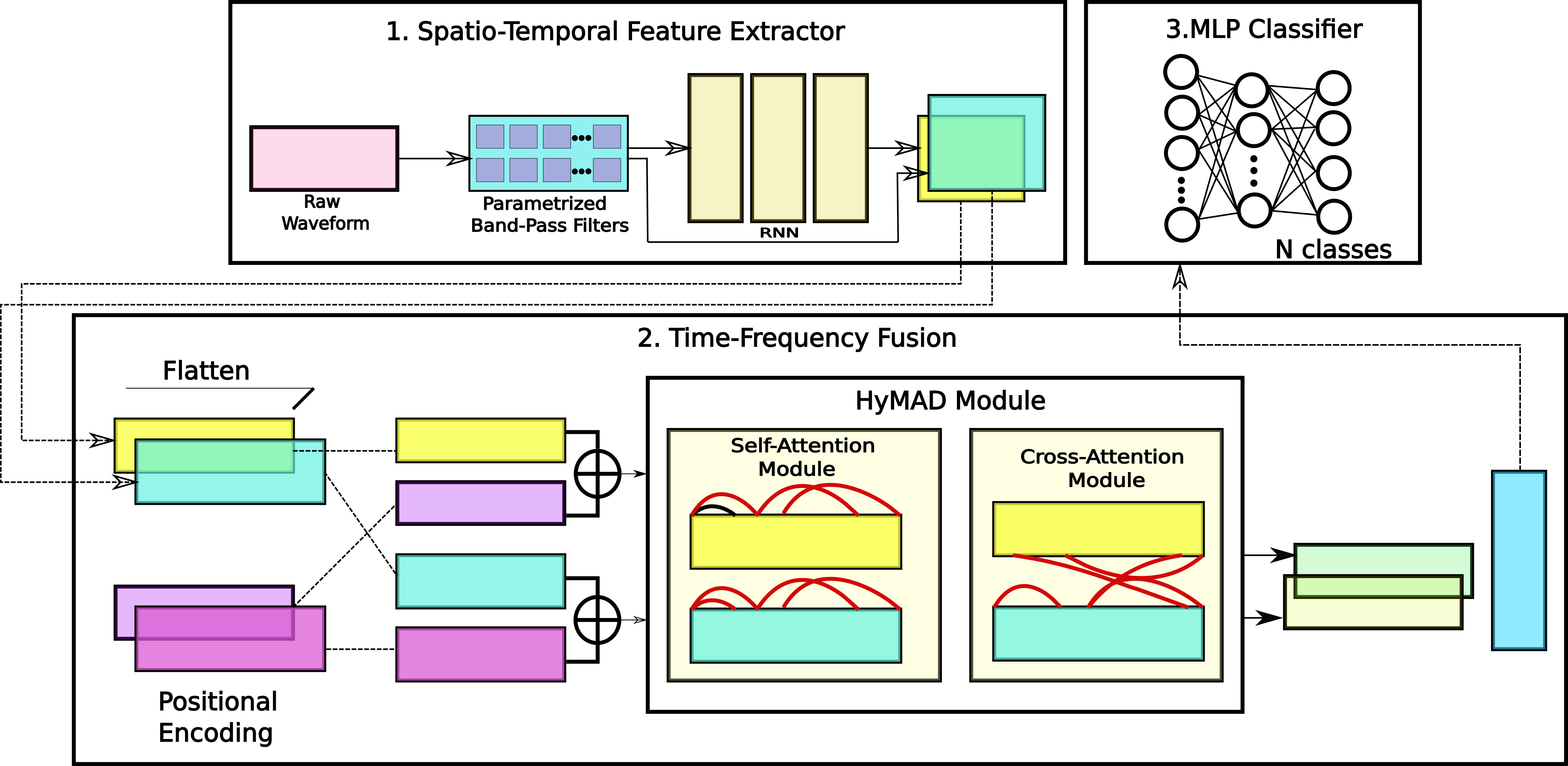}
    \caption{Overall architecture of the proposed Hybrid Multi-Activity Detection (HyMAD) model. The raw seismic signal is processed through parallel frequency and temporal feature extraction paths, which are then integrated via a cross-attention mechanism before classification.}
    \label{fig:hymad_architecture}
\end{figure*}

Given a seismic signal $\mathbf{x} \in \mathbb{R}^{1 \times T}$, existing approaches can identify single events occurring in isolated environments, but struggle to detect multiple overlapping events. To overcome this limitation, we propose \textbf{HyMAD} (Hybrid Multi-Activity Detection), an attention-based time-frequency fusion architecture that integrates a learnable frequency feature extractor, a temporal pattern encoder, and an attention-based fusion module to jointly model both spectral and temporal characteristics. An overview of the proposed method is presented in a flow diagram (Fig. \ref{fig:hymad_architecture}).

\subsection{Spatio-Temporal Feature Extractor}

Reliable detection of multiple events in a seismic signal depends on extracting both frequency and temporal features from the raw data. The spatio-temporal feature extractor serves as a foundational module in our architecture, designed to identify distinctive signal patterns that manifest across time and frequency domain within the seismic waveforms.

\subsubsection{Learnable Frequency Encoder using SincNet}

We employ a SincNet's~\cite{ravanelli2018speaker} parameterized convolution layer to extract robust and interpretable spectral features from raw seismic signals. Unlike traditional convolutional neural networks (CNNs)~\cite{bengio} that learn arbitrary filter shapes, SincNet utilizes parameterized convolutional filters that function as band-pass filters, making them highly suitable for spectral analysis of seismic data. Whereas a conventional band-pass filter isolates a specific frequency range in the frequency domain, in the time domain this operation can be approximated using a sinc function. SincNet leverages this property by expressing each filter \(h_k[n]\) as the difference between two low-pass sinc functions, thus forming a band-pass filter. Unlike hand-crafted frequency features, the lower and upper cutoff frequencies of these filters are directly parameterized and learned during training, enabling the model to adaptively select relevant frequency bands from the data.

For each filter \(k\) in the SincNet layer, the time-domain filter kernel is defined as \(h_k[n]\), where \(n\) is the sample index within the filter kernel, \(n = 0, 1, \ldots, L-1\), and \(L\) is the kernel length:
\begin{equation}
h_k[n] = 2f_{2,k}\,\mathrm{sinc}(2\pi f_{2,k} n) - 2f_{1,k}\,\mathrm{sinc}(2\pi f_{1,k} n)
\label{eq:sinc_filter}
\end{equation}
where \(\mathrm{sinc}(x) = \sin(x)/x\). Here, \(f_{1,k}\) and \(f_{2,k}\) are the learnable lower and upper cutoff frequencies for the \(k\)-th filter, with the constraint \(0 \leq f_{1,k} < f_{2,k}\). These cutoff frequencies are learned directly from data, allowing the model to adaptively identify the most salient frequency bands for activity detection.

The output feature map at time \(m\) for each filter \(k\) is obtained by convolving the input seismic signal \(x[n]\) with the learned filter \(h_k[n]\) as defined in Eq.~\ref{eq:sinc_filter}:
\begin{equation}
y_k[m] = (\mathbf{x} * h_k)[m] = \sum_{n=0}^{L-1} x[m-n] \cdot h_k[n]
\label{eq:conv}
\end{equation}

Collectively, the outputs from all \(C\) filters at all time indices form the SincNet feature map, as given by Eq.~\ref{eq:conv}:
\begin{equation}
\mathbf{Z}_{\text{sinc}} = \operatorname{SincNet}(\mathbf{x}) = [y_1, y_2, \ldots, y_C ] \in \mathbb{R}^{C \times L_1}
\label{eq:z_sinc}
\end{equation}
where \(\mathbf{x} \in \mathbb{R}^{1 \times T}\) is the input seismic signal of length \(T\), \(C\) is the number of filters, and \(L_{1}\) is the temporal length of the output after convolution (where \(L_1 = T\) when using same padding).

\subsubsection{Temporal Feature Encoding with RNN}

Following spectral feature extraction by the SincNet layer, the next step in our HyMAD architecture is to capture the inherent temporal dynamics and long-range dependencies present in these features. Seismic signals are inherently sequential, and activities often exhibit distinctive temporal patterns, durations, and overlapping sequences. While frequency-based analysis captures spectral content, it lacks the capacity to model its evolution over time. To address this, we employ a recurrent neural network (RNN)~\cite{ELMAN1990179} module for temporal feature encoding.

At each time step \(t\), the input to the RNN is a feature vector \(\mathbf{f}_t\) (derived from the SincNet output), characterizing the spectral content at that instant. The RNN maintains a hidden state \(\mathbf{h}_t\) that serves as a summary of information from previous time steps, thereby providing memory of past events.

Let \(\mathbf{Z}_{\text{sinc}} \in \mathbb{R}^{C \times T}\) denote the SincNet output, where \(C\) is the number of filters and \(T\) the temporal length. We represent this as a sequence \(\mathbf{F} = [\mathbf{f}_1, \mathbf{f}_2, \ldots, \mathbf{f}_T]\), with \(\mathbf{f}_t \in \mathbb{R}^C\) at each time \(t\).

\paragraph{RNN Computation.}
For each step \(t\), the hidden state is updated as:
\begin{equation}
\mathbf{h}_t = \phi\left( \mathbf{W}_h \mathbf{h}_{t-1} + \mathbf{W}_x \mathbf{f}_t + \mathbf{b} \right),
\label{eq:rnn_update}
\end{equation}
where \(\mathbf{h}_{t-1}\) is the previous hidden state, \(\mathbf{W}_h\) and \(\mathbf{W}_x\) are the hidden-to-hidden and input-to-hidden weight matrices, \(\mathbf{b}\) is a bias vector, and \(\phi(\cdot)\) denotes a non-linear activation function (typically \(\tanh\)).

The sequence of hidden states forms the RNN feature matrix:
\begin{equation}
\mathbf{H}_{\mathrm{RNN}} = [\mathbf{h}_1, \mathbf{h}_2, \ldots, \mathbf{h}_T] \in \mathbb{R}^{H \times T},
\label{eq:rnn_output}
\end{equation}
where \(H\) is the hidden state dimensionality.

Each \(\mathbf{h}_t \in \mathbb{R}^H\) encodes the temporal context up to time \(t\), providing a rich feature representation for downstream HyMAD modules. This recurrent mechanism allows the RNN to capture both short- and long-term dependencies across the SincNet-derived frequency feature sequence, as described in Eq.~\ref{eq:rnn_update}.

\subsection{Spatio-Temporal Fusion}

After spatial and temporal features are extracted, \(\mathbf{Z}_{\text{sinc}}\) and \(\mathbf{H}_{\mathrm{RNN}}\) are processed by the HyMAD fusion module to obtain position- and context-dependent local features. Intuitively, this module converts the features into representations that facilitate alignment and matching between temporal and spectral information. We denote these transformed features as \(\mathbf{Z}^{\mathrm{tr}}_{\text{sinc}}\) and \(\mathbf{H}^{\mathrm{tr}}_{\mathrm{RNN}}\).

\paragraph{Preliminaries: Transformer.} 
A Transformer encoder~\cite{vaswani2017attention} is composed of sequentially connected layers, each centered around an attention mechanism. The key component is the attention layer, where input vectors are named query (\(Q\)), key (\(K\)), and value (\(V\)). The query retrieves information from the value, weighted by its similarity (dot product) with the key:
\begin{equation}
\mathrm{Attention}(Q, K, V) = \mathrm{softmax}\left(\frac{Q K^\top}{\sqrt{d_k}}\right) V,
\label{eq:attention}
\end{equation}
where \(d_k\) is the key dimension. The attention operation enables elements in a sequence to attend to each other, facilitating message passing and global context aggregation. Equation~\ref{eq:attention} serves as the fundamental operation for both self- and cross-attention mechanisms in our fusion architecture.

\subsubsection{Positional Encoding}

In time--frequency fusion, maintaining temporal ordering is crucial, as the evolution and duration of events provide key characteristics for distinguishing different activities. However, because self-attention is permutation invariant, each feature sequence is augmented with positional information, enabling the model to reason about order and relative distances.

Let the input to the positional encoding module be the sequence:
\begin{equation}
\mathbf{E} = [\mathbf{e}_1, \mathbf{e}_2, \ldots, \mathbf{e}_T] \in \mathbb{R}^{T \times d_{\text{model}}},
\label{eq:input_sequence}
\end{equation}
where \(T\) is the temporal length and \(d_{\text{model}}\) the feature dimension.

The positional encoding function computes a deterministic matrix \(\mathbf{P} \in \mathbb{R}^{T \times d_{\text{model}}}\) as defined in Eqs.~\ref{eq:posenc_sin} and~\ref{eq:posenc_cos}:
\begin{align}
\mathbf{P}_{t,2i}     &= \sin \left( \frac{t}{10000^{2i/d_{\text{model}}}} \right), 
\label{eq:posenc_sin}\\
\mathbf{P}_{t,2i+1}   &= \cos \left( \frac{t}{10000^{2i/d_{\text{model}}}} \right),
\label{eq:posenc_cos}
\end{align}
for \(t = 0, \ldots, T-1\) and \(i = 0, \ldots, \lfloor d_{\text{model}}/2 \rfloor - 1\).

The positionally encoded sequence is obtained by summing the input and positional embeddings as in Eq.~\ref{eq:encoded_sequence}:
\begin{equation}
\widetilde{\mathbf{E}} = \mathbf{E} + \mathbf{P},
\label{eq:encoded_sequence}
\end{equation}
where \(\widetilde{\mathbf{E}} \in \mathbb{R}^{T \times d_{\text{model}}}\). This representation is then input to subsequent self-attention blocks governed by Eq.~\ref{eq:attention}.

Specifically, for both feature streams, the positionally encoded SincNet and RNN feature sequences are given by Eqs.~\ref{eq:posenc_sinc} and~\ref{eq:posenc_rnn}, respectively:
\begin{align}
\widetilde{\mathbf{F}}_{\text{sinc}}     &= \mathbf{F}_{\text{sinc}} + \mathbf{P}, 
\label{eq:posenc_sinc}\\
\widetilde{\mathbf{H}}_{\mathrm{RNN}}    &= \mathbf{H}_{\mathrm{RNN}} + \mathbf{P}.
\label{eq:posenc_rnn}
\end{align}

\subsubsection{Attention-Based Fusion via Self and Cross-Attention}

The positionally encoded sequences \(\widetilde{\mathbf{F}}_{\text{sinc}}\) and
\(\widetilde{\mathbf{H}}_{\mathrm{RNN}}\) are independently passed through
dedicated self-attention blocks (as in Eq.~\ref{eq:attention}), enabling
each modality to weigh the importance of elements within its own domain
and capture long-range intra-modal dependencies. The resulting refined
representations are denoted as:

\begin{align}
\mathbf{A}_{\text{sinc}} &= \mathrm{SelfAttn}\big(\widetilde{\mathbf{F}}_{\text{sinc}},
\widetilde{\mathbf{F}}_{\text{sinc}}, \widetilde{\mathbf{F}}_{\text{sinc}}\big),
\label{eq:selfattn_sinc}\\
\mathbf{A}_{\mathrm{RNN}} &= \mathrm{SelfAttn}\big(\widetilde{\mathbf{H}}_{\mathrm{RNN}},
\widetilde{\mathbf{H}}_{\mathrm{RNN}}, \widetilde{\mathbf{H}}_{\mathrm{RNN}}\big).
\label{eq:selfattn_rnn}
\end{align}

The refined embeddings \(\mathbf{A}_{\text{sinc}}\) and
\(\mathbf{A}_{\mathrm{RNN}}\) are then used in a bidirectional
cross-attention mechanism, allowing the frequency and temporal streams to
attend to one another and exchange complementary information. Specifically,
the two directional cross-attention operations are defined as:
\begin{align}
\mathbf{C}_{\text{freq}} &= \mathrm{CrossAttn}\big(
\mathbf{A}_{\text{sinc}}, \mathbf{A}_{\mathrm{RNN}}, \mathbf{A}_{\mathrm{RNN}} \big),
\label{eq:crossattn_freq}\\
\mathbf{C}_{\text{temp}} &= \mathrm{CrossAttn}\big(
\mathbf{A}_{\mathrm{RNN}}, \mathbf{A}_{\text{sinc}}, \mathbf{A}_{\text{sinc}} \big),
\label{eq:crossattn_temp}
\end{align}
where the first corresponds to the frequency-query attending to temporal
keys and values, and the second represents the temporal-query attending to
frequency features.

Finally, the cross-attended outputs are fused by concatenation to produce
a unified joint representation:
\begin{equation}
\mathbf{Z}_{\text{CA}} = \mathbf{C}_{\text{freq}} \oplus \mathbf{C}_{\text{temp}},
\label{eq:crossattn_fusion}
\end{equation}
where \([\cdot \oplus \cdot]\) denotes concatenation along the feature
dimension. The resulting embedding \(\mathbf{Z}_{\text{CA}}\) integrates
both spectral and temporal context and serves as the input to the final
classification module.

\subsection{MLP Classification Layer}

The fused feature representation $\mathbf{Z}_{\text{CA}}$, synthesizing both spectral and temporal characteristics via self- and cross-attention, serves as the input to the final multi-label classification module. The classifier is a multi-layer perceptron (MLP) with two fully connected layers, each followed by a non-linear ReLU activation. The final layer outputs one (unactivated) logit per class. During training, we use the \texttt{BCEWithLogitsLoss} criterion, which applies a sigmoid function to each output and computes the binary cross-entropy for each class independently.

Formally, the process is:
\begin{equation}
\begin{aligned}
\mathbf{h}^{(1)} &= \phi\left( \mathbf{Z}_{\mathrm{CA}} \mathbf{W}_1 + \mathbf{b}_1 \right) \\
\mathbf{h}^{(2)} &= \phi\left( \mathbf{h}^{(1)} \mathbf{W}_2 + \mathbf{b}_2 \right) \\
&\vdots \\
\mathbf{o} &= \mathbf{h}^{(L-1)} \mathbf{W}_L + \mathbf{b}_L
\end{aligned}
\end{equation}
where $\phi(\cdot)$ is the non-linear (ReLU) activation, and $L$ is the total number of layers. The output $\mathbf{o} \in \mathbb{R}^C$ represents the logits for each class.

In this way, the MLP serves as the final component of the HyMAD pipeline, transforming the fused learned features into reliable multi-label predictions for diverse and overlapping seismic events.

\section{Implementation Details}
The HyMAD architecture was implemented using the PyTorch deep learning framework (version 2.4.0) in Python (version 3.10.8), and experiments were conducted on a workstation equipped with an NVIDIA RTX 4070 (12 GB memory) and an Intel i9-13900HX processor with 16 GB RAM. Training was performed using the AdamW optimizer with a fixed learning rate of 1e-2, a batch size of 128, and for a maximum of 200 epochs; as this is a multi-label classification problem, the Binary Cross-Entropy with Logits Loss was used as the objective function throughout all experiments.

\section{Dataset}
\subsection{Dataset Acquisition}

\begin{figure}[!t]
    \centering
    \includegraphics[width=0.45\textwidth]{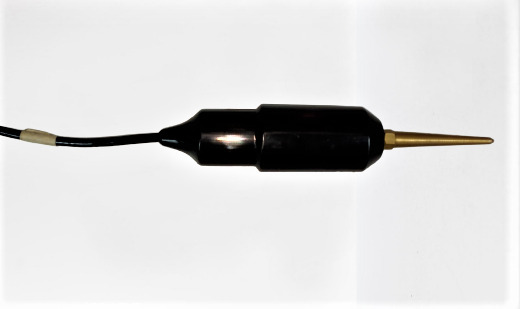}
    \caption{Geophone used in the experimental setup.}
    \label{fig:geophone}
\end{figure}

\begin{figure}[!t]
    \centering
    \includegraphics[width=0.45\textwidth]{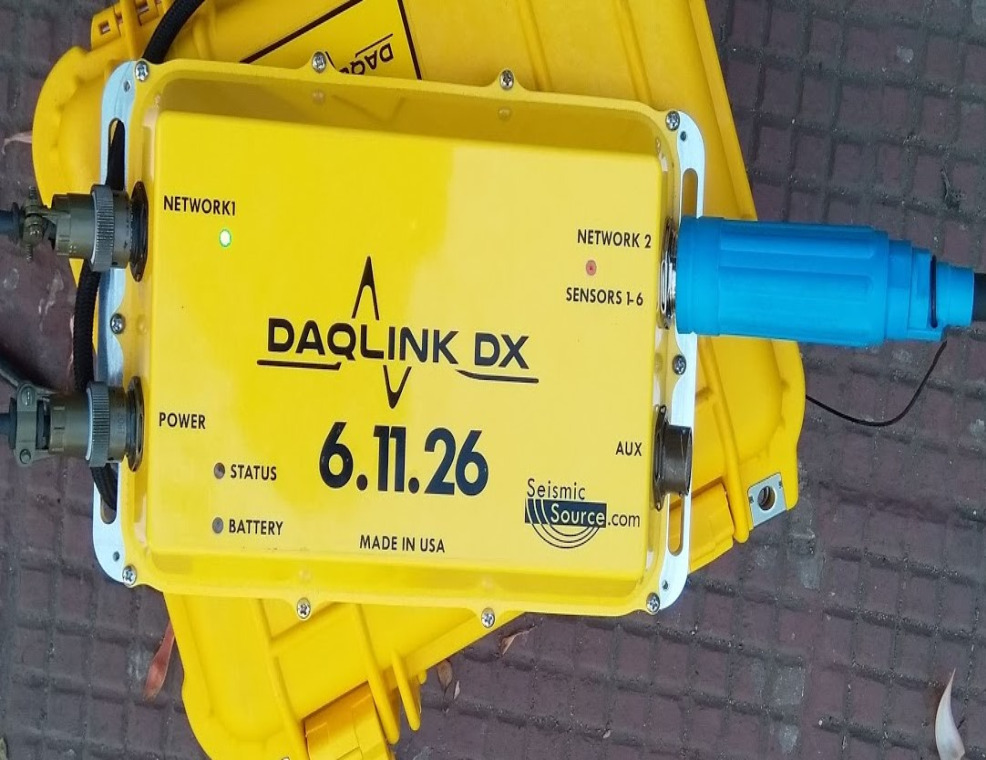}
    \caption{Data acquisition system used in the experimental setup.}
    \label{fig:daq}
\end{figure}

Data required for this experiment was collected using a geophone array deployed near the CSIR-CMERI Robotics \& Automation Division in Durgapur. The data is collected from the geophone at an 8 kHz sampling rate. During seismic occurrences, movement relative to the suspended spring coil and casing inside the geophone induces an output voltage. Using a 24-bit Data Acquisition System, this analog voltage data is converted into digital time series data. An array of six sensors spaced 15 meters apart covers a 100-meter perimeter; all six sensors are synchronously connected to the central Data Acquisition system. The details of the geophones and Data Acquisition system are listed in Table \ref{tab:geophone_spec} and Table \ref{tab:daq_spec}, respectively. The physical setup is further illustrated in Figure \ref{fig:geophone} and \ref{fig:daq}.

The collected dataset encompasses four distinct single-activity events, each acquired separately: human footstep movement, animal footstep movement, speeding vehicle movement, and periods of no movement at all (referred to as "No Event").
To prevent data leakage, this complete dataset of single-activity recordings was first rigorously divided into an 80\% training set, 10\% validation set, and a 10\% test set, ensuring no overlap between splits. Instances of multiple concurrent activities (e.g., human+animal, human+vehicle) were then generated only by combining signals from within their respective data splits (e.g., training signals were only merged with other training signals). For superposition, we adopted a realistic generation protocol to simulate the variability of field conditions by introducing two key random parameters. First, to simulate asynchronous activities, the secondary signal was shifted by a random time delay relative to the primary signal, ensuring the model learns to detect overlaps that do not necessarily start at the same instant. Additionally, to account for varying signal strengths, each signal was independently scaled by a random factor before superposition. This simulates varying distances of the sources from the sensor and different source energies, ensuring the model is robust to scenarios where one activity, such as a vehicle, might energetically dominate another, like a footstep. The final dataset, consisting of both the original single-activity signals and these generated multi-activity samples, was used for training and evaluation.

\vspace{0.5em}

\subsubsection{Generation of Overlapping Signals}
To simulate realistic scenarios where multiple activities occur simultaneously, the overlapping signals in our dataset were generated by superimposing individual activity signals. This approach involves summing the time-domain waveforms of two or more distinct single-activity events (e.g., a human footstep signal and an animal footstep signal) that were recorded separately.

This method is a widely accepted and correct approach for generating synthetic multi-source signals in seismic research for the following reasons:
\begin{enumerate}
    \item \textbf{Physics-Based Linear Superposition:} For the low-amplitude vibrations induced by border surveillance targets (e.g., human footsteps, light vehicles), the soil medium operates primarily within its linear elastic regime. Unlike high-magnitude events (e.g., earthquakes) where non-linear soil deformation occurs, surveillance targets generate micro-seismic strains where the stress-strain relationship remains linear. Consequently, the principle of superposition holds valid, meaning the total displacement field generated by multiple concurrent sources is mathematically equivalent to the linear sum of the displacement fields produced by each source individually.
        \item \textbf{Controlled Ground Truth:} Generating overlapping signals synthetically provides precise control over the ground truth. We know exactly which individual activities are present and at what time instances, which is invaluable for training and rigorously evaluating a multi-label detection model. Collecting naturally occurring multi-activity seismic data with accurate, frame-level ground truth is extremely challenging and often impractical.
    \item \textbf{Focus on Disentanglement:} This approach allows us to isolate and specifically test the model's ability to disentangle blended signatures. If the model can accurately identify superimposed signals, it demonstrates its capacity to learn the unique underlying features of each activity even when they are combined, which is the core challenge of multi-activity detection.
    \item \textbf{Resemblance to Real-World Concurrency:} While collected individually, the superposition process effectively creates a signal that resembles the physical phenomenon of multiple sources vibrating the ground simultaneously. The resulting waveform captures the combined spectral and temporal characteristics that would be observed when activities like a human walking and a vehicle passing occur at the same time near a sensor..\vspace{0.5em}
\end{enumerate}
This synthetic generation strategy allows for a robust and controlled experimental environment to develop and validate our HyMAD architecture's capability in identifying complex, concurrent seismic events.

\subsection{Dataset Characteristics and Pre-processing}
Each data sample in our dataset corresponds to a fixed-duration segment of \textbf{1} seconds. Given the 8 kHz sampling rate, each segment contains \textbf{8000} data points.

The dataset is structured to represent a multi-label classification problem. While each of the four primary single-activity classes (human, animal, vehicle, no event) contains 4000 raw data samples, the superimposed multi-activity classes (human+animal, human+vehicle, vehicle+animal) also consist of 4000 samples each, ensuring a comprehensive representation of concurrent events. The distribution of these multi-activity combinations is designed to reflect varying degrees of overlap and complexity.

Prior to being fed into the HyMAD architecture, raw seismic signals undergo minimal pre-processing to preserve their inherent characteristics. Each segment is first normalized to a standard range (e.g., zero mean and unit variance) to ensure consistent input scales across all samples. This normalization helps in stabilizing model training and preventing dominance by signals with higher amplitudes. No explicit filtering or advanced noise reduction techniques are applied at this stage, as the architecture is designed to learn robust features directly from the potentially noisy raw data. The 8 kHz sampling rate provides sufficient temporal resolution for capturing the nuances of various seismic events.

\begin{table}[!t]
\centering
\caption{Specification of Geo-phone}
\label{tab:geophone_spec}
\begin{tabular}{ll}
\toprule
\textbf{Specification} & \textbf{Value} \\
\midrule
Make/Model        & HSG / HG-24 U 10 Hz @375$\Omega$ \\
Natural Frequency & 10 Hz $\pm$ 2.5\% \\
Sensitivity (V/(m/s)) & 20.95 $\pm$ 2.5\% \\
String Resistance & 273$\Omega$ $\pm$ 2.5\% \\
Harmonic Distortion & $<$0.1\% \\
\bottomrule
\end{tabular}
\end{table}

\begin{table}[!t]
\centering
\caption{Specification of Data Acquisition System}
\label{tab:daq_spec}
\begin{tabular}{ll}
\toprule
\textbf{Specification} & \textbf{Value} \\
\midrule
Make/Model      & Seismic Source HGS/Daqlink DX6 \\
Resolution      & 24 bits \\
Interface       & TCP/IP \\
Pre-amp Gain    & $\times$1, $\times$4, $\times$16 \\
Sampling Rate   & Multiples of 125 Sps \\
Bandwidth       & DC to 20 KHz \\
Input Impedance & 100 K$\Omega$ \\
\bottomrule
\end{tabular}
\end{table}




\section{Experiments}

\subsection{Baseline Approaches}

We evaluate our proposed model against a range of standard baselines. The classical machine learning models include Random Forests \cite{breiman2001random}, Extra Trees \cite{geurts2006extremely}, XGBoost \cite{chen2016xgboost}, LightGBM \cite{ke2017lightgbm}, CatBoost \cite{prokhorenkova2018catboost}, and Gaussian Mixture Models \cite{reynolds2009gmm} (GMM). In addition, we also compare with deep learning architectures such as a Multi-Layer Perceptron (MLP) \cite{rumelhart1986learning} and a Long Short-Term Memory (LSTM) \cite{hochreiter1997long} network. Each model was trained with commonly adopted hyperparameter configurations to ensure fair comparison (see Table~\ref{tab:model_params}). Ensemble tree-based methods were trained with 200 estimators and shallow depths to avoid overfitting. Boosting frameworks such as XGBoost, LightGBM, and CatBoost were run in CPU mode with parallel execution enabled. Neural models (MLP and LSTM) were trained for 200 epochs with Adam optimizer and batch size of 128.

\begin{table*}[ht]
\centering
\caption{Training setup for baseline models.}
\label{tab:model_params}
\begin{tabular}{p{0.2\textwidth} p{0.75\textwidth}}
\toprule
\textbf{Model} & \textbf{Training Setup} \\
\midrule
RandomForest   & Ensemble of 200 decision trees, trained with parallel processing, fixed seed of 42. \\
ExtraTrees     & Ensemble of 200 extremely randomized trees, trained with parallel processing, fixed seed of 42. \\
GradientBoosting & Boosting with 200 trees, shallow depth, moderate learning rate, fixed seed of 42. \\
XGBoost        & Histogram-based boosting, trained on CPU/GPU with log-loss as evaluation metric, parallel execution. \\
LightGBM       & Gradient boosting framework trained on CPU with default parameters and parallel execution. \\
CatBoost       & Gradient boosting on CPU, silent mode, with automatic handling of categorical features. \\
SVM-RBF        & Support Vector Machine with RBF kernel, probability estimates enabled, standard regularization setup. \\
GMM            & Gaussian Mixture Model with two components, diagonal covariance matrices, regularized estimation. \\
LSTM (PyTorch) & Single-layer LSTM with hidden size 64, optimized using Adam, trained for 200 epochs with batch size 128. \\
MLP (PyTorch)  & Three-layer feedforward network (256 -- 128 -- output), ReLU activations, optimized with Adam, trained for 200 epochs with batch size 128. \\
\bottomrule
\end{tabular}
\end{table*}

\medskip
\noindent\textbf{Evaluation Metrics.} 
To assess the performance of all models, we employ multiple evaluation metrics. Strict Match Accuracy (also called exact match ratio) requires the predicted label set 
for each instance to exactly match the ground truth label set, making it a highly stringent metric. In contrast, Hamming Accuracy evaluates correctness at the per-label level and allows partial credit when some but not all labels are predicted correctly. 
These metrics are reported alongside the standard measures of Precision, Recall, F1-score, and AUROC (see Table~\ref{tab:results}).

\emph{Strict Match Accuracy} (also called exact match ratio) requires that the predicted label set for each instance exactly matches the true label set, making it a highly stringent measure. Even a single label mismatch counts the entire prediction as incorrect. Formally, 
$\text{SMA} = \tfrac{1}{N}\sum_{i=1}^N \mathbf{1}\{Y_i = \hat{Y}_i\}$, 
where $Y_i$ is the ground-truth label set and $\hat{Y}_i$ is the predicted label set. 

On the other hand, \emph{Hamming Accuracy} evaluates correctness at the per-label level and thus allows partial credit when only some labels are predicted correctly. It is defined as 
$\text{HA} = 1 - \tfrac{1}{N \cdot L}\sum_{i=1}^N \sum_{j=1}^L \mathbf{1}\{Y_{ij} \neq \hat{Y}_{ij}\}$, 
where $L$ is the number of labels and $Y_{ij}, \hat{Y}_{ij}$ are the true and predicted values for the $j$-th label of instance $i$.

\medskip
\noindent\textbf{Feature Extraction.} 
For spectral signal representation, we consider three feature extraction methods: Mel-Frequency Cepstral Coefficients (MFCC), Log-Mel Spectrogram, and Low-Frequency Spectrum Cepstral Coefficients (LFSCC). These features capture different aspects of the spectral domain, providing compact and discriminative representations of the underlying signal. The parameters used for feature extraction are summarized in Table~\ref{tab:feature_params}.

\begin{table}[t]
\centering
\caption{Feature extraction parameters for MFCC, Log-Mel Spectrogram, and LFSCC.}
\label{tab:feature_params}
\resizebox{\columnwidth}{!}{%
\begin{tabular}{lccc}
\toprule
\textbf{Parameter} & \textbf{MFCC} & \textbf{Log-Mel} & \textbf{LFSCC} \\
\midrule
Sample Rate        & 8000 Hz & 8000 Hz & 8000 Hz \\
FFT Size ($n_{fft}$) & 256 & 256 & 256 \\
Window Length      & 256 & 256 & 256 \\
Hop Length         & 128 & 128 & 128 \\
Number of Filters  & 40 Mel & 40 Mel & Low 1/4 Spectrum \\
Number of Coeffs   & 13 & -- & 13 (DCT) \\
Feature Domain     & Cepstral (time--freq) & Frequency (Mel scale) & Low-frequency Cepstral \\
Normalization      & Mean-variance & Log-power & DCT-based \\
\bottomrule
\end{tabular}%
}
\end{table}

\medskip
\noindent\textbf{One-vs-Rest Classifier.} 
To evaluate classical machine learning approaches in a multi-label setting, we employ the OneVsRestClassifier strategy from \texttt{scikit-learn}. Most classical models in \texttt{scikit-learn} (e.g., SVM, Random Forest, Gradient Boosting) do not natively support multi-label classification. Hence, we use the \texttt{OneVsRestClassifier()} wrapper, which enables such models 
to operate in multi-label tasks. In this framework, a separate binary classifier is trained for each label, treating the presence of that label as the positive class and all other labels 
as the negative class. During inference, each classifier outputs the probability of its corresponding label being present, and the final multi-label prediction is obtained by aggregating the results across all labels. 

For example, when using XGBoost with four target labels, the framework trains four independent models, one for each label, where each model performs binary classification. However, these classifiers operate independently and fail to account for correlations or co-occurrences among labels. In other words, the presence of one signal in the data does not influence how another label is predicted. In contrast, our proposed approach processes all features jointly within a unified representation space, enabling the model to capture inter-label dependencies and contextual relationships between signals in a single forward pass. With LFSCC features, GMM failed to model the multi-modal distribution entirely, predicting No Event for all complex cases.

\begin{table*}[ht]
\centering
\caption{Performance of baseline models across MFCC, Log-Mel Spectrogram, and LFSCC features. Metrics include strict accuracy (Exact Match), per-label accuracy (Hamming), Precision, Recall, F1-score, and AUROC. The proposed HyMAD model outperforms or matches baselines across most metrics.}
\label{tab:results}
\resizebox{\textwidth}{!}{%
\begin{tabular}{llllllll}
\toprule
\textbf{Model} & \textbf{Feature} & \textbf{Exact Match Acc} & \textbf{Hamming Acc} & \textbf{Precision} & \textbf{Recall} & \textbf{F1} & \textbf{AUROC} \\
\midrule
RandomForest   & MFCC   & 0.810 & 0.871 & 0.825 & 0.785 & 0.805 & 0.963 \\
ExtraTrees     & MFCC   & 0.830 & 0.883 & 0.835 & 0.824 & 0.828 & 0.969 \\
XGBoost        & MFCC   & 0.852 & 0.891 & 0.920 & 0.766 & 0.836 & 0.983 \\
LightGBM       & MFCC   & 0.841 & 0.872 & 0.902 & 0.766 & 0.828 & 0.974 \\
CatBoost       & MFCC   & 0.848 & 0.878 & 0.917 & 0.766 & 0.835 & 0.983 \\
GMM            & MFCC   & 0.000 & 0.649 & 0.500 & 0.486 & 0.492 & 0.597 \\
LSTM (Torch)   & MFCC   & 0.798 & 0.929 & 0.907 & 0.911 & 0.909 & 0.973 \\
MLP (Torch)    & MFCC   & 0.801 & 0.864 & 0.747 & 0.918 & 0.821 & 0.971 \\
\midrule
RandomForest   & LogMel & 0.812 & 0.858 & 0.758 & 0.912 & 0.830 & 0.963 \\
ExtraTrees     & LogMel & 0.831 & 0.865 & 0.878 & 0.766 & 0.818 & 0.974 \\
XGBoost        & LogMel & 0.876 & 0.921 & 0.967 & 0.766 & 0.855 & 0.973 \\
LightGBM       & LogMel & 0.850 & 0.912 & 0.924 & 0.761 & 0.838 & 0.985 \\
CatBoost       & LogMel & 0.832 & 0.894 & 0.881 & 0.766 & 0.827 & 0.975 \\
GMM            & LogMel & 0.000 & 0.649 & 0.500 & 0.486 & 0.492 & 0.597 \\
LSTM (Torch)   & LogMel & 0.656 & 0.899 & 0.708 & 0.636 & 0.664 & 0.951 \\
MLP (Torch)    & LogMel & 0.833 & 0.919 & 0.859 & 0.897 & 0.866 & 0.970 \\
\midrule
RandomForest   & LFSCC  & 0.735 & 0.916 & 0.852 & 0.940 & 0.894 & 0.974 \\
ExtraTrees     & LFSCC  & 0.715 & 0.911 & 0.840 & 0.941 & 0.887 & 0.971 \\
XGBoost        & LFSCC  & 0.767 & 0.921 & 0.876 & 0.920 & 0.897 & 0.974 \\
LightGBM       & LFSCC  & 0.758 & 0.918 & 0.867 & 0.923 & 0.894 & 0.975 \\
CatBoost       & LFSCC  & 0.785 & 0.926 & 0.879 & 0.929 & 0.903 & 0.977 \\
GMM            & LFSCC  & 0.000 & 0.649 & 0.500 & 0.486 & 0.492 & 0.597 \\
LSTM (Torch)   & LFSCC  & 0.137 & 0.756 & 0.550 & 0.402 & 0.465 & 0.768 \\
MLP (Torch)    & LFSCC  & 0.467 & 0.781 & 0.847 & 0.589 & 0.559 & 0.655 \\
\midrule
\rowcolor{gray!20}
\textbf{HyMAD (Proposed)} & \textbf{Custom} & \textbf{0.904} & \textbf{0.974} & \textbf{0.980} & \textbf{0.962} & \textbf{0.964} & \textbf{0.995} \\
\bottomrule
\end{tabular}%
}
\end{table*}

\subsection{Ablation Studies}

To quantify the contribution of key components in the proposed HyMAD
framework, we perform a series of ablation experiments by selectively
removing or modifying critical modules. Table~\ref{tab:ablation}
summarizes the performance degradation observed under each configuration.

\textbf{(a) Without Temporal Modeling.} In this variant, the recurrent
temporal modeling module is removed, and classification is performed
directly on frame-level spectral features. The performance drops sharply
to 58.79\% accuracy and 73.57\% precision, indicating that temporal
dependencies are crucial for correctly recognizing sustained or
transitional events such as vehicle motion.

\textbf{(b) Without Multi-Scale Frequency Extraction.} Here, the
multi-scale SincNet front-end is replaced with a single-scale convolution.
The model achieves 62.45\% accuracy and 70.32\% precision, confirming
that multi-scale representations are essential for capturing the
fine-grained spectral variations necessary to distinguish between
activities such as footsteps and animal movements.

\textbf{(c) Naïve Fusion (Concat).} The cross-attention fusion block is
replaced with a simple concatenation of temporal and spectral embeddings.
Although this variant performs better than the single-stream configurations,
achieving 91.26\% accuracy and 90.32\% precision, it still underperforms
the proposed cross-attention fusion mechanism, demonstrating that naive
feature aggregation fails to fully capture inter-modal dependencies.

\textbf{(d) HyMAD (Proposed).} The complete model, integrating multi-scale
frequency extraction, recurrent temporal modeling, and bidirectional
cross-attention fusion, achieves the highest performance across all
metrics, demonstrating the complementary nature of each component.

\begin{table}[h!]
\centering
\caption{Performance comparison of different model variants.}
\label{tab:ablation}
\setlength{\tabcolsep}{4pt} 
\renewcommand{\arraystretch}{1.1} 
\scriptsize 
\begin{tabular}{lcccc}
\toprule
\textbf{Model} & \textbf{F1 (\%)} & \textbf{Precision (\%)} & \textbf{Recall (\%)} & \textbf{AUROC} \\
\midrule
w/o temporal modeling   & 58.79 & 73.57 & 48.96 & 82.01 \\
w/o Multi-Scale features  & 62.45 & 70.32 & 54.23 & 85.65 \\
Naive fusion (Concat)   & 91.26 & 90.32 & 87.65 & 97.65 \\
\textbf{HyMAD (Proposed)} & \textbf{95.28} & \textbf{96.81} & \textbf{93.94} & \textbf{99.45} \\
\bottomrule
\end{tabular}
\end{table}

\subsection{Visualization and Analysis}

\subsubsection{t-SNE Visualization}
\begin{figure}[h!]
\centering
\includegraphics[width=0.45\textwidth]{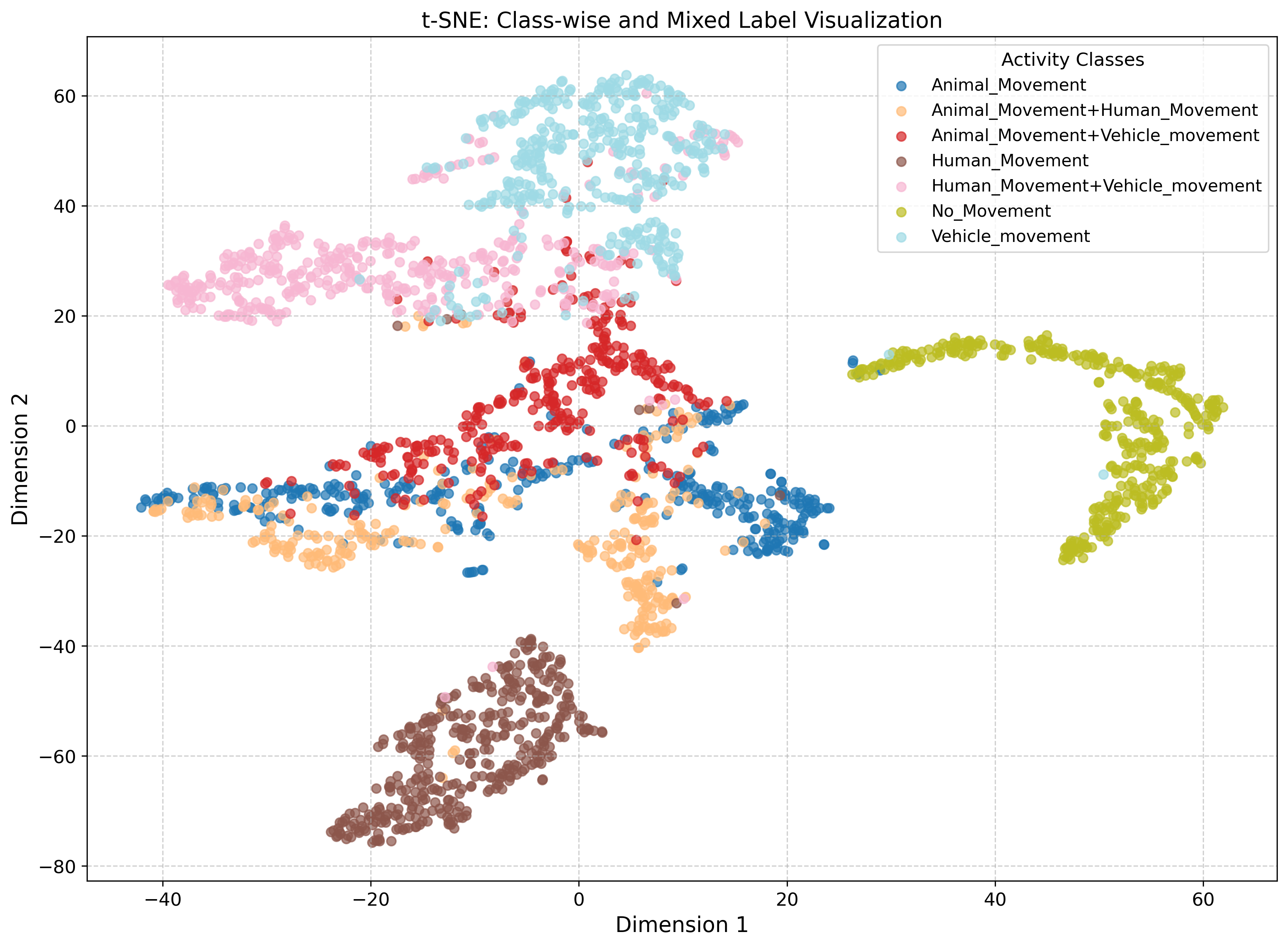}
\caption{t-SNE visualization of feature embeddings.}
\label{fig:tsne}
\end{figure}

To gain insights into the feature representations learned by our
model, we apply t-SNE to the embeddings of the penultimate
layer. As shown in Fig.~\ref{fig:tsne}, the features form
well-separated clusters corresponding to different classes,
demonstrating the model’s ability to learn discriminative
representations even in the presence of overlapping signals.

\subsubsection{ROC Curve Analysis}
\begin{figure}[h!]
    \centering
    \begin{subfigure}{0.48\linewidth}
        \centering
        \includegraphics[width=\linewidth]{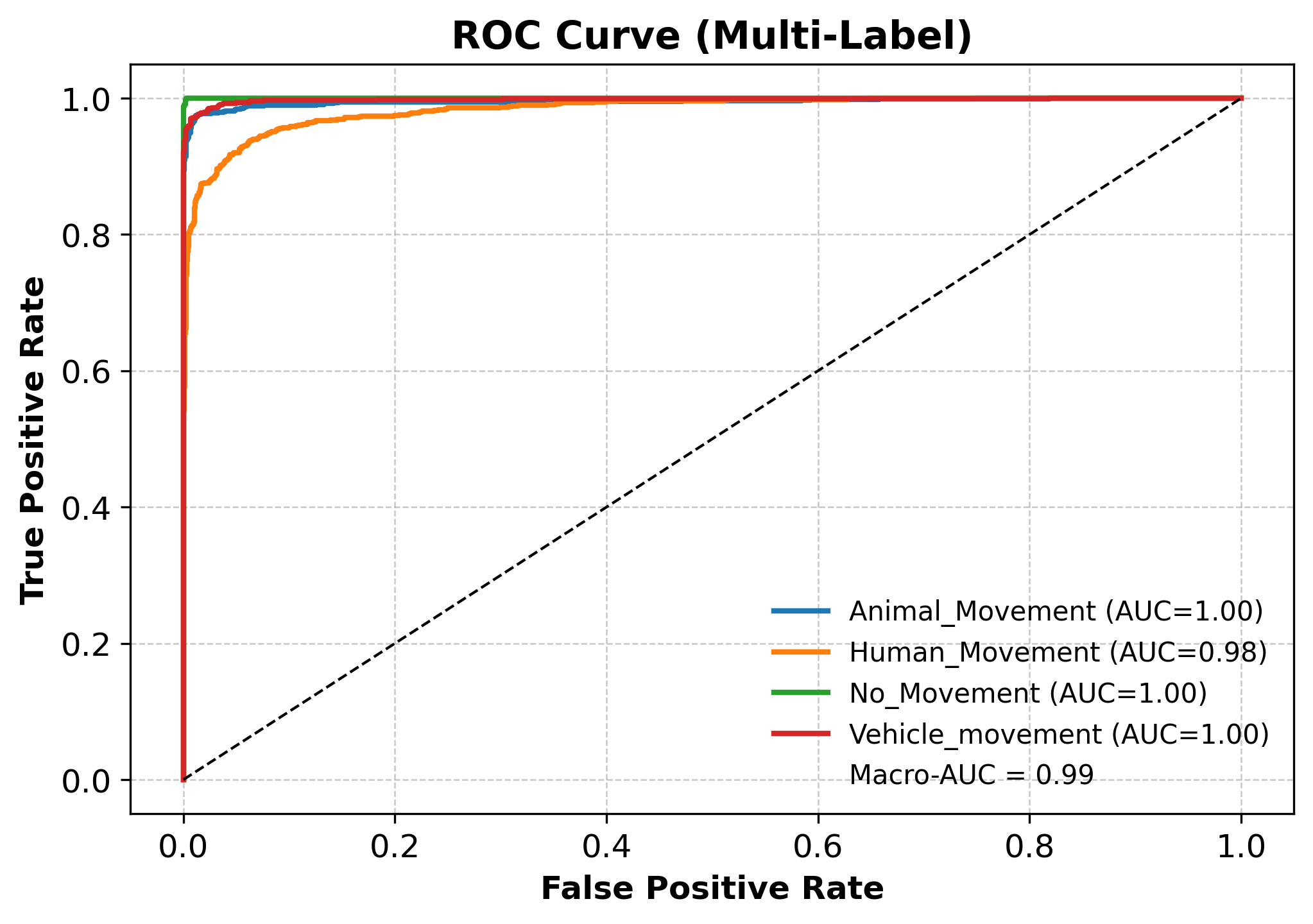}
        \caption{ROC curve}
        \label{fig:roc}
    \end{subfigure}
    \hfill
    \begin{subfigure}{0.48\linewidth}
        \centering
        \includegraphics[width=\linewidth]{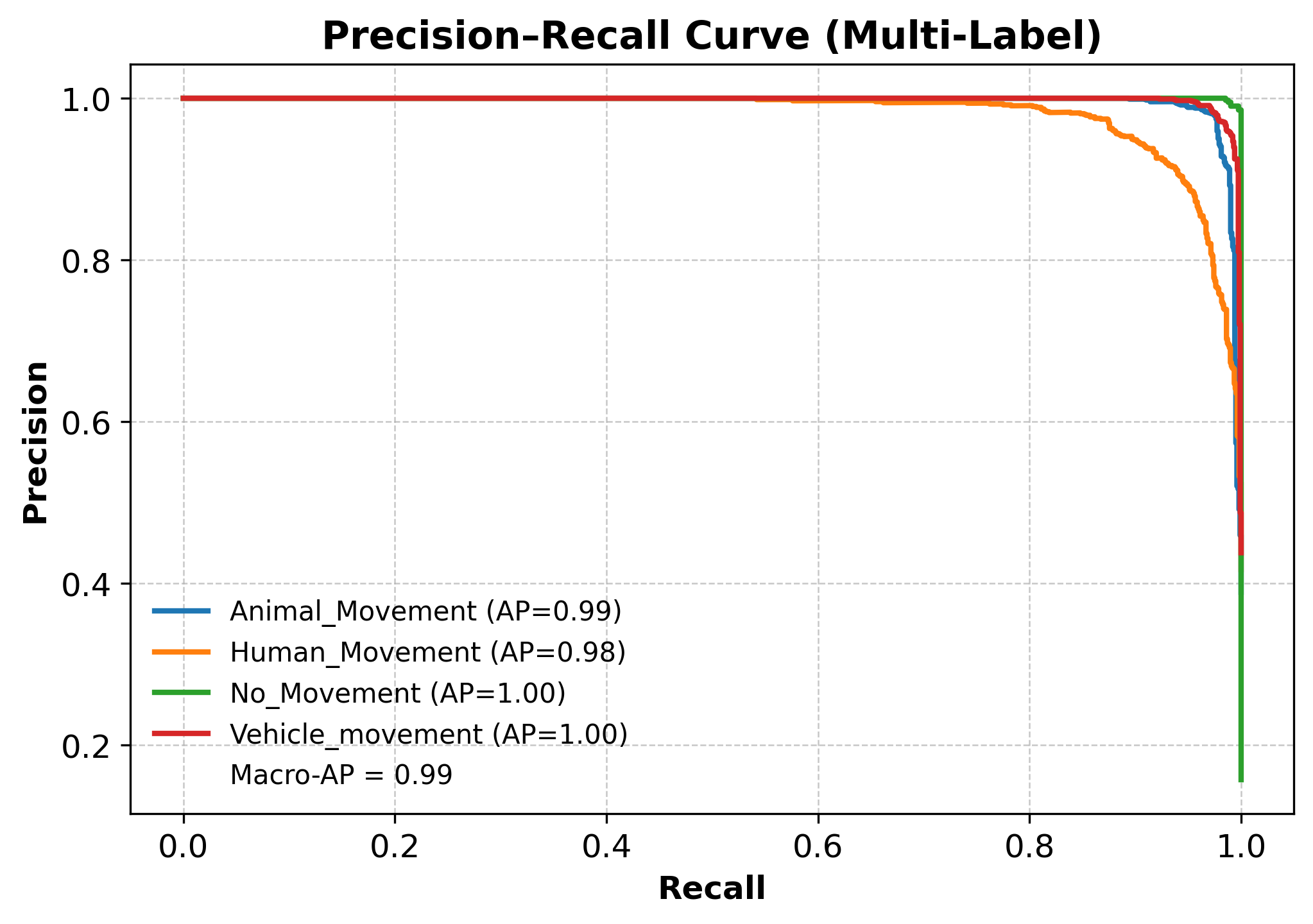}
        \caption{PR curve}
        \label{fig:prc}
    \end{subfigure}
    \caption{Comparison of ROC (left) and PR (right) curves for the proposed HyMAD model.}
    \label{fig:roc_pr_horizontal}
\end{figure}

We further evaluate the discriminative ability of our model 
using ROC curves. Fig.~\ref{fig:roc} shows the ROC plots for
different classes, with our method achieving high AUC scores 
across all categories. This demonstrates the robustness of the
proposed model under varying signal conditions.


\section{Conclusion}

In this work, we introduced HyMAD (Hybrid Multi-Activity Detection), a novel deep learning framework for the challenging task of multi-label seismic event classification. Our approach addresses the significant challenge of disentangling overlapping activities, such as human footfalls, animal movements, and vehicle passages, from complex and noisy seismic signals. By leveraging a multi-representation fusion strategy, HyMAD effectively combines learnable frequency-oriented features, extracted via a SincNet-based convolutional layer, with temporal dependencies modeled by a recurrent neural network (RNN). The integration of dedicated self-attention blocks for intra-modal contextualization and a cross-attention fusion module proved crucial for adaptively combining these complementary features and enabling the robust identification of simultaneous events.

The results of our evaluation on a real-world surveillance dataset demonstrate that HyMAD not only achieves competitive performance but also exhibits a strong ability to generalize to complex, concurrently occurring activity scenarios. The proposed modular architecture provides a robust solution for real-world security applications, overcoming the limitations of traditional methods that struggle with the intricate and overlapping patterns inherent in multi-source seismic data. Our work represents a significant step forward in the field of seismic signal analysis, offering a principled and effective framework for advancing real-time activity recognition and enhancing the effectiveness of seismic-based monitoring systems in challenging environments.

\section{Limitation and Future Work}

Although the proposed HyMAD model shows strong results and robustness against varying signal strengths and time offsets, this study has a limitation that can guide future work. The primary constraint is that the overlapping events are generated via the physics-based superposition of isolated field recordings. While this method provides precise ground truth and allows for controlled testing of disentanglement capabilities, it effectively simulates the linear combination of signals. In rare, high-energy scenarios or specific soil conditions, non-linear interactions between ground waves might occur which are not fully captured by linear superposition.

Future work will focus on deploying the HyMAD system in a continuous, real-time border surveillance pilot to collect and evaluate performance on naturally occurring, concurrent multi-source events. Additionally, we aim to explore the integration of unsupervised domain adaptation techniques to further improve the model's generalization across diverse geological terrains without requiring extensive re-training.


\bibliographystyle{IEEEtran}
\bibliography{references}

\begin{thebibliography}{10}
\providecommand{\url}[1]{#1}
\csname url@samestyle\endcsname
\providecommand{\newblock}{\relax}
\providecommand{\bibinfo}[2]{#2}
\providecommand{\BIBentrySTDinterwordspacing}{\spaceskip=0pt\relax}
\providecommand{\BIBentryALTinterwordstretchfactor}{4}
\providecommand{\BIBentryALTinterwordspacing}{\spaceskip=\fontdimen2\font plus
\BIBentryALTinterwordstretchfactor\fontdimen3\font minus \fontdimen4\font\relax}
\providecommand{\BIBforeignlanguage}[2]{{%
\expandafter\ifx\csname l@#1\endcsname\relax
\typeout{** WARNING: IEEEtran.bst: No hyphenation pattern has been}%
\typeout{** loaded for the language `#1'. Using the pattern for}%
\typeout{** the default language instead.}%
\else
\language=\csname l@#1\endcsname
\fi
#2}}
\providecommand{\BIBdecl}{\relax}
\BIBdecl

\bibitem{10630594}
S.~Aruchamy, A.~Chakraborty, M.~Das, S.~R.~K. Vadali, R.~Ray, and S.~Nandy, ``An efficient {Gaussian} mixture model classifier for outdoor surveillance using seismic signals,'' \emph{IEEE Transactions on Geoscience and Remote Sensing}, vol.~62, pp. 1--11, 2024.

\bibitem{HASABNIS2025105815}
P.~Hasabnis, E.~A. Nilot, and Y.~E. Li, ``Introducing {USED}: Urban seismic event detection,'' \emph{Computers \& Geosciences}, vol. 196, p. 105815, 2025.

\bibitem{yoon2017overlap}
C.~Yoon, O.~O'Reilly, K.~Bergen, and G.~C. Beroza, ``Detection and classification of seismic events with overlapping waveforms,'' \emph{Science Advances}, vol.~3, no.~6, p. e1700578, 2017.

\bibitem{yilmaz2001seismic}
Ã.~Yilmaz, \emph{Seismic Data Analysis}.\hskip 1em plus 0.5em minus 0.4em\relax Society of Exploration Geophysicists, 2001.

\bibitem{1164317}
D.~Griffin and J.~Lim, ``Signal estimation from modified short-time {Fourier} transform,'' \emph{IEEE Transactions on Acoustics, Speech, and Signal Processing}, vol.~32, no.~2, pp. 236--243, 1984.

\bibitem{morlet1982wave}
J.~Morlet, G.~Arens, E.~Fourgeau, and D.~Giard, ``Wave propagation and sampling theory—part {I}: Complex signal and scattering in multilayered media,'' \emph{Geophysics}, vol.~47, no.~2, pp. 203--221, 1982.

\bibitem{cortes1995support}
C.~Cortes and V.~Vapnik, ``Support-vector networks,'' \emph{Machine Learning}, vol.~20, no.~3, pp. 273--297, 1995.

\bibitem{breiman2001random}
L.~Breiman, ``Random forests,'' \emph{Machine Learning}, vol.~45, no.~1, pp. 5--32, 2001.

\bibitem{rabiner1989tutorial}
L.~R. Rabiner, ``A tutorial on hidden {Markov} models and selected applications in speech recognition,'' \emph{Proceedings of the IEEE}, vol.~77, no.~2, pp. 257--286, 1989.

\bibitem{lecun1998gradient}
Y.~LeCun, L.~Bottou, Y.~Bengio, and P.~Haffner, ``Gradient-based learning applied to document recognition,'' \emph{Proceedings of the IEEE}, vol.~86, no.~11, pp. 2278--2324, 1998.

\bibitem{elman1990finding}
J.~L. Elman, ``Finding structure in time,'' \emph{Cognitive Science}, vol.~14, no.~2, pp. 179--211, 1990.

\bibitem{hochreiter1997long}
S.~Hochreiter and J.~Schmidhuber, ``Long short-term memory,'' \emph{Neural Computation}, vol.~9, no.~8, pp. 1735--1780, 1997.

\bibitem{cho2014learning}
K.~Cho, B.~van Merri{\"e}nboer, C.~Gulcehre, D.~Bahdanau, F.~Bougares, H.~Schwenk, and Y.~Bengio, ``Learning phrase representations using {RNN} encoder-decoder for statistical machine translation,'' in \emph{Proceedings of the 2014 Conference on Empirical Methods in Natural Language Processing (EMNLP)}, 2014, pp. 1724--1734.

\bibitem{ravanelli2018speaker}
M.~Ravanelli and Y.~Bengio, ``Speaker recognition from raw waveform with {SincNet},'' in \emph{Proc. Interspeech}, 2018, pp. 1956--1960.

\bibitem{chang2021mssincresnetjointlearning1d}
\BIBentryALTinterwordspacing
P.-C. Chang, Y.-S. Chen, and C.-H. Lee, ``{MS-SincResNet}: Joint learning of {1D} and {2D} kernels using multi-scale {SincNet} and {ResNet} for music genre classification,'' 2021. [Online]. Available: \url{https://arxiv.org/abs/2109.08910}
\BIBentrySTDinterwordspacing

\bibitem{mayortorres2021interpretablesincnetbaseddeeplearning}
\BIBentryALTinterwordspacing
J.~M. Mayor-Torres, M.~Ravanelli, S.~E. Medina-DeVilliers, M.~D. Lerner, and G.~Riccardi, ``Interpretable {SincNet}-based deep learning for emotion recognition from {EEG} brain activity,'' 2021. [Online]. Available: \url{https://arxiv.org/abs/2107.10790}
\BIBentrySTDinterwordspacing

\bibitem{Luo2023}
J.~Luo, J.~Li, Q.~Mao, Z.~Shi, H.~Liu, X.~Ren, and X.~Hei, ``Overlapping filter bank convolutional neural network for multisubject multicategory motor imagery brain-computer interface,'' \emph{BioData Mining}, vol.~16, no.~1, p.~19, 2023.

\bibitem{Chen2023}
Z.~Chen, D.~Gao, K.~Sun, X.~Zhao, Y.~Yu, and Z.~Wang, ``Densely connected networks with multiple features for classifying sound signals with reverberation,'' \emph{Sensors}, vol.~23, no.~16, p. 7225, 2023.

\bibitem{Fedorishin2021}
\BIBentryALTinterwordspacing
D.~Fedorishin, N.~Sankaran, D.~D. Mohan, J.~Birgiolas, P.~Schneider, S.~Setlur, and V.~Govindaraju, ``Waveforms and spectrograms: Enhancing acoustic scene classification using multimodal feature fusion,'' in \emph{DCASE Workshop 2021}, 2021, pp. 216--220. [Online]. Available: \url{https://dcase.community/documents/workshop2021/proceedings/DCASE2021Workshop_Fedorishin_69.pdf}
\BIBentrySTDinterwordspacing

\bibitem{McLoughlin2020}
I.~McLoughlin, Z.~Xie, Y.~Song, H.~Phan, and R.~Palaniappan, ``Time-frequency feature fusion for noise robust audio event classification,'' \emph{Circuits, Systems, and Signal Processing}, vol.~39, no.~3, pp. 1672--1687, 2020.

\bibitem{Wang2022}
B.~Wang, G.~Chen, L.~Rong, Y.~Liu, A.~Yu, X.~He, T.~Wen, Y.~Zhang, and B.~Hu, ``Arrhythmia disease diagnosis based on {ECG} time-frequency domain fusion and convolutional neural network,'' \emph{IEEE Journal of Translational Engineering in Health and Medicine}, vol.~11, pp. 116--125, 2022.

\bibitem{Jin2022}
Y.~Jin, M.~Wang, L.~Luo, D.~Zhao, and Z.~Liu, ``Polyphonic sound event detection using temporal-frequency attention and feature space attention,'' \emph{Sensors}, vol.~22, no.~18, p. 6818, 2022.

\bibitem{202412.1180}
R.~Huang, X.~Yue, and J.~Pengxu, ``Local time-frequency feature fusion using cross-attention for acoustic scene classification,'' \emph{Preprints}, 2024.

\bibitem{bengio}
Y.~Bengio and Y.~Lecun, ``Convolutional networks for images, speech, and time-series,'' 1997.

\bibitem{ELMAN1990179}
J.~L. Elman, ``Finding structure in time,'' \emph{Cognitive Science}, vol.~14, no.~2, pp. 179--211, 1990.

\bibitem{vaswani2017attention}
A.~Vaswani, N.~Shazeer, N.~Parmar, J.~Uszkoreit, L.~Jones, A.~N. Gomez, L.~Kaiser, and I.~Polosukhin, ``Attention is all you need,'' in \emph{Advances in Neural Information Processing Systems}, 2017, pp. 5998--6008.

\bibitem{geurts2006extremely}
P.~Geurts, D.~Ernst, and L.~Wehenkel, ``Extremely randomized trees,'' \emph{Machine Learning}, vol.~63, no.~1, pp. 3--42, 2006.

\bibitem{chen2016xgboost}
T.~Chen and C.~Guestrin, ``{XGBoost}: A scalable tree boosting system,'' in \emph{Proceedings of the 22nd ACM SIGKDD International Conference on Knowledge Discovery and Data Mining}, 2016, pp. 785--794.

\bibitem{ke2017lightgbm}
G.~Ke, Q.~Meng, T.~Finley, T.~Wang, W.~Chen, W.~Ma, Q.~Ye, and T.-Y. Liu, ``{LightGBM}: A highly efficient gradient boosting decision tree,'' in \emph{Advances in Neural Information Processing Systems}, 2017, pp. 3146--3154.

\bibitem{prokhorenkova2018catboost}
L.~Prokhorenkova, G.~Gusev, A.~Vorobev, A.~V. Dorogush, and A.~Gulin, ``{CatBoost}: unbiased boosting with categorical features,'' in \emph{Advances in Neural Information Processing Systems}, 2018, pp. 6638--6648.

\bibitem{reynolds2009gmm}
D.~A. Reynolds, ``{Gaussian} mixture models,'' MIT Lincoln Laboratory, Tech. Rep., 2009.

\bibitem{rumelhart1986learning}
D.~E. Rumelhart, G.~E. Hinton, and R.~J. Williams, ``Learning representations by back-propagating errors,'' \emph{Nature}, vol. 323, pp. 533--536, 1986.

\end{thebibliography}

\end{document}